\documentclass[10pt,letterpaper]{article}
\usepackage[top=0.85in,left=2.75in,footskip=0.75in]{geometry}

\usepackage{amsmath,amssymb}

\usepackage{changepage}

\usepackage{textcomp,marvosym}

\usepackage{cite}

\usepackage{nameref,hyperref}

\usepackage[right]{lineno}

\usepackage[nopatch=eqnum]{microtype}
\DisableLigatures[f]{encoding = *, family = * }

\usepackage[table, dvipsnames]{xcolor}

\usepackage{array}

\usepackage{longtable}

\usepackage{hyperref}

\newcolumntype{+}{!{\vrule width 2pt}}

\newlength\savedwidth

\raggedright
\usepackage[aboveskip=1pt,labelfont=bf,labelsep=period,justification=raggedright,singlelinecheck=off]{caption}

\makeatletter
\renewcommand{\@biblabel}[1]{\quad#1.}
\makeatother

\usepackage{lastpage,fancyhdr,graphicx}
\usepackage{epstopdf}

\fancyheadoffset[L]{2.25in}
\fancyfootoffset[L]{2.25in}
\usepackage{booktabs}
\usepackage{multirow}
\usepackage{siunitx}

\begin{document}
\vspace*{0.2in}

\begin{flushleft}
{\Large
\textbf\newline{Data Augmentation of Time-Series Data in Human Movement Biomechanics: A~Scoping Review} 
}
\newline
\\
Christina Halmich\textsuperscript{1,2*},
Lucas Höschler\textsuperscript{3\Yinyang},
Christoph Schranz\textsuperscript{2\Yinyang},
Christian Borgelt\textsuperscript{1},
\\
\bigskip
\textbf{1} Department of Artificial Intelligence and Human Interfaces, Paris-Lodron-University Salzburg, Salzburg, Austria
\\
\textbf{2} Salzburg Research Forschungsgesellschaft mbH, Jakob-Haringer-Straße 5/3, 5020 Salzburg, Austria
\\
\textbf{3} Department of Sport and Exercise Science,  Paris-Lodron-University Salzburg, Salzburg, Austria
\\
\bigskip

%
%
\Yinyang These authors contributed equally to this work.





* christina.halmich@salzburgresearch.at

\end{flushleft}

\textbf{This is a preprint version of a manuscript currently under review at PLOS ONE. The content may change during the peer review process.}

\section*{Abstract}
\textbf{BACKGROUND:} The integration of machine learning and deep learning methodologies has transformed data analytics in biomechanics, supported by extensive wearable sensor data. However, the field faces challenges such as limited large-scale data sets and high data acquisition costs that hinder the development of robust algorithms. Data augmentation techniques have shown promise in addressing these issues, yet their application in biomechanical time-series data requires comprehensive evaluation.
\newline
\textbf{OBJECTIVE:} This study conducts a scoping review of research on data augmentation for time-series data in the biomechanics domain. It focuses on analyzing current methods used for augmenting and generating time-series data sets, evaluates the effectiveness of these methods, and gives recommendations for working with augmentation methods in the biomechanics domain. 
\newline
\textbf{DESIGN:} Four online databases - PubMed, IEEE~Xplore, Scopus, and Web of Science - were used to find studies published between 2013 and 2024. Following the PRISMA-ScR guidelines, two screening processes were conducted to identify publications focused on the research topic of data augmentation for biomechanical time-series data.
\newline
\textbf{RESULTS:}
 After screening, 21 publications were identified as relevant to this topic. The results indicate that there is no single best practice method for augmenting biomechanical time-series data; instead, various methods are employed based on study aims. This review highlights the issue of synthetic data lacking soft tissue artifacts, leading to discrepancies known as the “synthetic gap”, emphasizing the need for more realistic data augmentation techniques. Furthermore, the evaluation of augmentation methods often lacks proper analysis, making it difficult to understand the effects of data created using different techniques. This understanding is crucial for assessing the impact of the augmented data set on downstream models and evaluating the quality of the data augmentation process.
\newline
\textbf{CONCLUSIONS:} This scoping review emphasizes the critical role of data augmentation in overcoming limited data set availability, enhancing model performance, and generalization in biomechanical time-series data. By tailoring augmentation techniques to data characteristics, future research can significantly advance the accuracy and applicability of predictive models in biomechanics. However, it is crucial to understand the impacts of data generated using different augmentation methods, as this understanding directly impacts the further development of improved methods.


\section*{Introduction}
With the rise and advancements of Machine Learning and Deep Learning methodologies, there has been a noticeable transformation in data analytics across various fields, including human movement biomechanics~\cite{xiang_recent_2022}.
The availability of wearable sensor data in recent years has further fueled this shift by enabling data-driven analytics through the generation of extensive data sets~\cite{camomilla_trends_2018}. However, the complex multi-variable nature of human movement data poses challenges for traditional analytical methods, necessitating approaches capable of handling data-intensive tasks~\cite{halilaj_machine_2018}.
Deep learning methods have emerged as a solution to address these challenges by facilitating the analyses of large biomechanical data sets, extraction of relevant features, uncovering hidden relationships, and revealing emerging trends, thus advancing the understanding of human movement dynamics~\cite{halilaj_machine_2018}. Despite the increasing adoption of deep learning in human movement biomechanics, the scarcity of large-scale biomechanical data sets remains a significant obstacle~\cite{knudson_confidence_2017}.
The acquisition of biomechanical data comes with particularly high effort, as the measurement setup requires physical effort from participants, extended measurement duration under varying influence factors, and time-consuming data preprocessing.
This makes many research projects economically challenging both in terms of cost and time~\cite{knudson_confidence_2017,eOptimalControlSimulation2019}. 
Additionally, the limited availability of participants, coupled with challenges in obtaining informed consent, is further underlining this issue~\cite{knudson_confidence_2017}. Moreover, data collection using sensors can be impeded by uncontrollable conditions (e.g., equipment malfunctions, human error, sensor errors due to soft tissue) leading to data quality issues or loss~\cite{dorschkyCNNBasedEstimationSagittal2020a}.
The scarcity of large data sets not only impedes the development of robust models but also results in data sets lacking diversity and representation, making it challenging to train models that generalize well to new data~\cite{fister_synthetic_2022}. To address this data sparsity, data augmentation techniques have emerged as a promising solution. Data augmentation involves expanding existing data sets and increasing data diversity by modifying original samples, thus enhancing the robustness and generalization capabilities of machine learning models~\cite{fister_synthetic_2022}. It is not uncommon for data generation and data augmentation to be used interchangeably. However, it is important to note that data generation involves the creation of new synthetic samples, whereas data augmentation pertains to the enhancement of existing data samples. For the sake of simplicity within this work, data augmentation will be taken to encompass data generation methods. 
Initially applied in image recognition, data augmentation techniques have proven effective in expanding data sets and creating diversity~\cite{iglesias_data_2023, shortenSurveyImageData2019, iwanaEmpiricalSurveyData2021}. However, the application of these techniques to time-series data, prevalent in human movement biomechanics, presents unique challenges. Time-series data augmentation requires careful consideration of temporal dependencies and domain-specific constraints to generate synthetic samples that accurately represent real-world phenomena~\cite{iglesias_data_2023, fister_synthetic_2022, iwanaEmpiricalSurveyData2021}.
Moreover, the aim of data synthesis in biomechanics is to generate data that not only looks visually realistic but also improves the quality of subsequent analysis such as classification tasks.

Existing augmentation techniques in biomechanics primarily rely on physics- or statistics-based methods, which may not represent valid biomechanical samples due to simplifications in formulas used or realism of rotations that are introduced~\cite{dorschkyCNNBasedEstimationSagittal2020a, eguchiInsoleBasedEstimationVertical2019, thAccelerationMagnitudeImpact2020}. Alternatively, data-driven methods utilizing machine learning techniques, such as Generative Adversarial Networks (GANs), diffusion models, or auto-regression, are not yet commonly used in biomechanics~\cite{liuSpatiotemporalKinematicCharacteristics2022, bicerGenerativeDeepLearning2022a} nor validated in the biomechanical setting. To the best of our knowledge, there exists no comprehensive comparison of various augmentation techniques nor clear guidance on which methods to employ. This gap in the literature can lead to inefficiencies and inconsistencies in research, as well as hinder advancements in the field. By systematically exploring and analyzing current data augmentation techniques used for biomechanical time-series data across various tasks, this work aims to provide a systematic overview of existing data augmentation methods and to offer guidance for researchers working with biomechanical time-series data.
Improvements in augmentation techniques in these areas are crucial because only with sufficient and well-augmented data can we enhance the accuracy and robustness of predictive models, facilitate the development of personalized training and rehabilitation programs, and ultimately lead to better health outcomes and performance optimizations.

\section*{Methods}
\subsection*{Eligibility Criteria}
Aligned with the PRISMA-ScR (Preferred Reporting Items for Systematic Reviews and Meta-Analyses Extension for Scoping Reviews~\cite{tricco2018prisma}) framework, this scoping review systematically searches, selects, analyzes, and reports relevant literature. Thereby ensuring transparency throughout the reviewing process. The eligibility criteria were set to ensure a focused and comprehensive review of studies and research concerning data augmentation for biomechanical time-series data in human movement biomechanics. The following criteria were established:
\begin{enumerate}
    \item \textbf{Publication Types:}
    \begin{itemize}
        \item Peer-reviewed journal articles 
        \item Conference proceedings
    \end{itemize}
    
    \item \textbf{Publication Period:}
    \begin{itemize}
        \item Publications between January 2013 and July 2024.
    \end{itemize}
    
    \item \textbf{Language:}
    \begin{itemize}
        \item Publications must be written in English.
    \end{itemize}
    
    \item \textbf{Topic Relevance:}
    \begin{itemize}
        \item Publications must address biomechanical aspects using data augmentation techniques on time-series data.
    \end{itemize}
    
    \item \textbf{Exclusion Criteria:}
    \begin{itemize}
        \item Publications that solely address biomechanical aspects without discussing data augmentation techniques.
        \item Studies that only utilize non-time-series data.
        \item Publications that do not analyze human movement, such as a focus on animal biomechanics or non-biological systems.
    \end{itemize}
\end{enumerate}

These criteria ensure that the review focuses specifically on relevant studies that contribute to the understanding of data augmentation techniques in the context of biomechanical time-series data analysis within human movement biomechanics.

\subsection*{Information Sources}
Four databases --- PubMed, Scopus, IEEE~Xplore, and Web of Science --- were searched for peer-reviewed journals and conference proceedings. While Scopus and Web of Science cover interdisciplinary research across various fields, IEEE~Xplore added conference proceedings from engineering, technology, and computer science. Additionally, PubMed was incorporated to encompass biomedical, life science, and health-related disciplines. Publications from January 2013 until July 2024 were considered. The initial database search was conducted in November 2023, extracting publications from PubMed, Scopus, and IEEE~Xplore. The decision to include Web of Science was made in January 2024. The final database search for all databases was conducted in July 20204.

\subsection*{Search}

Our search strategy was designed to identify all indexed publications utilizing time-series data generation methods within the biomechanics domain.

\subsubsection*{Initial Query Formulation}
To formulate effective search queries, we began by defining primary terms related to our topic. Key terms such as "time-series," "data augmentation," and "biomechanics" were selected to ensure the inclusion of papers that utilized data augmentation for biomechanical time-series data sets. We then explored synonyms and variations of these terms to achieve comprehensive coverage.
To capture the temporal aspects, we used multiple variants of the keywords including "time-series," "temporal," "sequential," "waveform," and "one-dimensional." The domain-specific terms were centered around biomechanics.
Keywords such as "data augmentation" and its synonyms like "synthetic*" and "generate*" were employed to filter for papers focused on data augmentation and synthetic data generation. The term "generate" produced numerous false positives, necessitating additional qualifiers like "data" and "sample" to improve specificity. Additionally, domain-specific features indicative of time-series data, such as "gait" and "kinematics," were incorporated.

\subsubsection*{Iterative Refinement Process}
Our search strategy underwent iterative refinement based on initial search results and feedback from collaborators. After an initial database search, we reviewed the retrieved articles and identified areas for refinement or expansion of our search terms. Adjustments were made to incorporate additional keywords, synonyms, and database-specific search techniques, enhancing the thoroughness of our approach. This iterative process systematically improved the effectiveness of our search strategy, ensuring comprehensive coverage of relevant literature.
\subsubsection*{Initial Search Challenges} 
The initial search yielded only a few relevant publications. Identifying two key papers that were missing from our results highlighted the need to explore alternative search terms commonly used in biomechanics. It was discovered that machine learning terms are infrequently used in biomechanical literature; instead, verbs denoting the outcome of the method, such as "synthetic data set" or "generated data," are preferred.

The term "data augmentation" yielded limited results, leading to the inclusion of synonyms and qualifiers. Similarly, alternative mechanical features indicative of time-series data were incorporated. To ensure relevance to machine learning, the term "learning" was refined to "machine learning" and "deep learning" due to initial challenges.

\subsubsection*{Final Search Queries} 
\label{sec:final_queries}
This procedure resulted in the formulation of four final search queries, where an asterisk is a wildcard operator: 
\begin{enumerate}
    \item ({\color{teal}"time-serie*"} OR {\color{teal}"timeserie*"} OR {\color{teal}"time serie*"} OR {\color{teal}"temporal data"} OR {\color{teal}"temporal sequence"} OR {\color{teal}"periodical data"} OR {\color{teal}"sequential data"} OR {\color{teal}"time structured"} OR {\color{teal}"time sequence"}) AND ({\color{teal}"data augmentation"} OR ({\color{teal}"synth*"} AND {\color{teal}"data"}) OR {\color{teal}"data generation"} OR {\color{teal}"data enhancement"} OR {\color{teal}"data enrichment"} OR {\color{teal}"data creation"}) AND {\color{teal}"biomech*"} AND ({\color{teal}"machine learning"} OR {\color{teal}"deep learning"})
    \item ({\color{teal}"time-serie*"} OR {\color{teal}"timeserie*"} OR {\color{teal}"time serie*"} OR {\color{teal}"temporal data"} OR {\color{teal}"temporal sequence"} OR {\color{teal}"periodical data"} OR {\color{teal}"sequential data"} OR {\color{teal}"time structured"} OR {\color{teal}"time sequence"}) AND (("synth*"OR {\color{teal}"generated"} OR {\color{teal}"augmented"} OR {\color{teal}"enhanced"} OR {\color{teal}"enriched"} OR {\color{teal}"created"}) AND ({\color{teal}"data"} OR {\color{teal}"sample*"}))  AND {\color{teal}"biomech*"} AND ({\color{teal}"machine learning"} OR {\color{teal}"deep learning"})  
    \item (({\color{teal}"synth*"} OR {\color{teal}"generat*"} OR {\color{teal}"augment*"} OR {\color{teal}"enhance*"} OR {\color{teal}"enrich*"} OR {\color{teal}"creat*"} OR {\color{teal}"simulat*"}) AND ({\color{teal}"data"} OR {\color{teal}"sample*"})) AND {\color{teal}"biomech*"} AND ({\color{teal}"machine learning"} OR {\color{teal}"deep learning"}) AND ( {\color{teal}"gait"} OR {\color{teal}"inertial sensors"} OR {\color{teal}"IMU"} OR {\color{teal}"kinematics"} OR {\color{teal}" Ground Reaction Force"} OR {\color{teal}"inverse dynamics"} OR {\color{teal}"joint moments"} OR {\color{teal}"kinetics"} OR {\color{teal}"waveform"} OR {\color{teal}"joint angles"} OR {\color{teal}"power"} OR {\color{teal}"one dimensional data"})
    \item ({\color{teal}"time-series"} OR {\color{teal}"timeseries"} OR {\color{teal}"time series"} OR {\color{teal}"temporal data"} OR {\color{teal}"temporal sequence"} OR {\color{teal}"periodical data"} OR {\color{teal}"sequential data"} OR {\color{teal}"time structured"} OR {\color{teal}"time sequence"}) AND (({\color{teal}"synth*"} OR {\color{teal}"generat*"} OR {\color{teal}"augment*"} OR {\color{teal}"enhance*"} OR {\color{teal}"enrich*"} OR {\color{teal}"creat*"} OR {\color{teal}"simulat*"}) AND ({\color{teal}"data"} OR {\color{teal}"sample*"}))AND {\color{teal}"biomech*"} AND ({\color{teal}"machine learning"} OR {\color{teal}"deep learning"})
\end{enumerate}

\noindent
Table~\ref{table1} shows the final queries and the number of publications found in each database. 
\begin{table} [!h]
  \caption{Number of publications found in each database with the corresponding search terms that include duplicates.}
  \begin{tabular}[!ht]{m{3cm}SSSS}
    \toprule
    \multirow{2}{*}{\textbf{Search Term}} &
      \multicolumn{4}{c}{\textbf{Number of Publications found}} \\
      & {IEEE~Xplore} & {Scopus} & {PubMed}  & {Web of Science}\\
      \midrule
  \textbf{1}  & 2 & 3 & 1 & 1  \\ \hline
     \textbf{2} & 4 & 10 & 7 & 10 \\ \hline
     \textbf{3} & 0 & 240 & 165 & 185 \\ \hline
    \textbf{4} & 19 & 30 & 14 & 19 \\
    \bottomrule
  \end{tabular}
\label{table1}
\end{table}

\subsubsection*{Validation}
To validate the effectiveness of our search strategy, we compared the retrieved search results to a set of known relevant articles identified from preliminary literature searches. This validation process ensured that our search strategy captured a comprehensive range of relevant literature. Additionally, we consulted with domain experts in biomechanics and data augmentation to review the search strategy and provide feedback on its comprehensiveness and relevance. By incorporating these validation steps, we enhanced the credibility of our methodology and ensured the robustness of our search strategy.

\subsection*{Selection of Sources of Evidence}
\begin{figure}[!h]
\includegraphics[width=10cm]{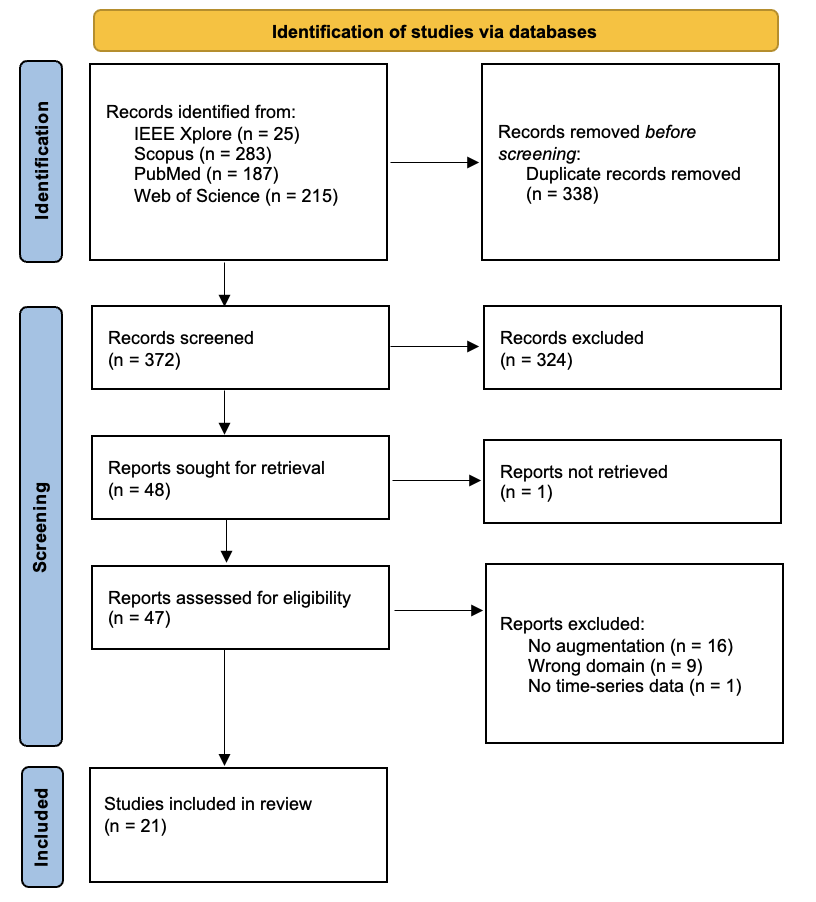}
\caption{{\bf Flow diagram of the systematic selection}
PRISMA 2020 flow diagram for systematic reviews based on the publications on this topic.}
\label{fig_prisma}
\end{figure}

Using the queries listed in the previous section~\nameref{sec:final_queries}, a total of 710 publications were found, published between January 2013 and July 2024. After identifying these publications, the next steps in the PRISMA guidelines to ensure a transparent and systematic selection process are screening and assessing the eligibility of publications to comprehensively capture relevant literature for our scoping review. Fig~\ref{fig_prisma} provides a visual summary of the systematic selection process, adhering to PRISMA guidelines for scoping reviews. After removing duplicates, 372 unique publications remained. 

\subsubsection*{First Screening Process}
In the initial screening, each publication's title and abstract were independently reviewed by at least two researchers to assess relevance to this scoping review. The criteria for relevance included:
\begin{itemize}
    \item Alignment with the scope of biomechanical time-series data augmentation in human movement
    \item Discussion of data augmentation methods
    \item Explicit aims to augment data or proposed methods suitable for data augmentation
\end{itemize}

This inclusion criterion ensured that only publications with data augmentation as the primary goal or relevant methodologies specifically designed for data augmentation were considered.
Any discrepancies or uncertainties during the screening were resolved through discussion and consensus among all researchers. This rigorous process ensured the quality and relevance of the included publications. 

\subsubsection*{Full-Text Analysis}
After the initial screening, 48~publications were identified as fitting the scope of this scoping review and were selected for full-text analysis. However, one publication was not available online and was excluded after the authors did not respond to our inquiries, leaving 47~publications for full-text review. Each of the 47~publications underwent a thorough review by at least two researchers. In cases of uncertainty, all researchers collectively examined and discussed the publication in question. During this process, the main data items explained in Section~\nameref{sec:data_items} were extracted.

\subsubsection*{Final Selection}
At the end of the detailed review process, 21~publications were included in the scoping review. This selection was based on the rigorous application of inclusion criteria and the detailed examination of each publication, ensuring the comprehensiveness and relevance of the included evidence. Table~\hyperref[S1_Table]{S1} presents the final 21~publications, including their titles, and authors.

\subsection*{Data Items} \label{sec:data_items}
To systematically and consistently capture the relevant information from the publications, the researchers completed a survey addressing various aspects of the studies. This survey facilitated the extraction of essential details from the publications, including the year and location of publication, specifics about the data used (such as type and sensor placement), the disciplinary context, and details about machine learning models trained on the augmented data set. Furthermore, the survey included an analysis of research gaps that necessitated the use of data augmentation techniques. The primary focus of the analysis was identifying the type of augmentation methods employed and the methodologies used to generate synthetic data in the reviewed publications.

\newpage
\section*{Results}
\subsection*{Geographic and Temporal Trends}
\begin{figure}[!h]
\includegraphics[width=10cm]{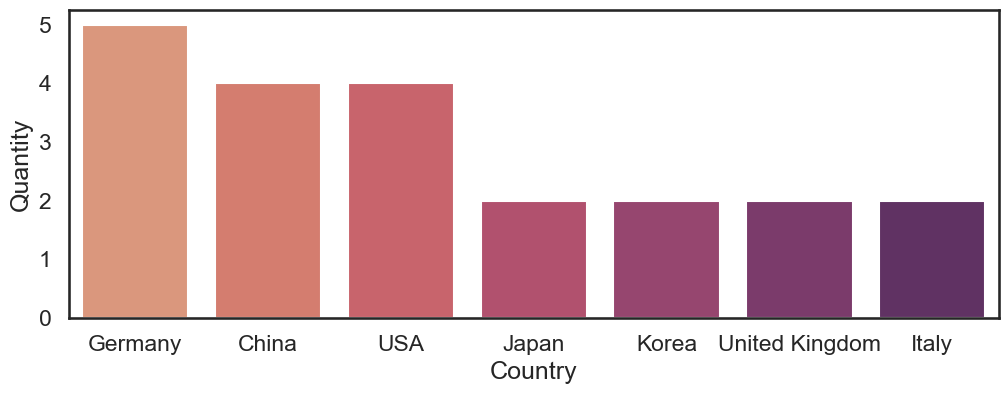}
\caption{{\bf Geographical Distribution of Publications}
Geographical distribution of the publications extracted from the affiliation of the first author and ranked by the number of publications.}
\label{fig_countries}
\end{figure}

The publications show a diverse geographical distribution, as illustrated in Fig~\ref{fig_countries}. The data was extracted based on the affiliation of the corresponding first author. Germany was the most prolific contributor with five publications~\cite{eOptimalControlSimulation2019,mPredictionLowerLimb2020,zimmermannIMUtosegmentAssignmentOrientation2018,dorschkyCNNBasedEstimationSagittal2020a,mundtEstimationGaitMechanics2020}. China and the United States followed closely with four publications~\cite{jProbabilityFusionApproach2023,liuSpatiotemporalKinematicCharacteristics2022,liangEstimationElectricalMuscle2024,tangSyntheticIMUDatasets2024},
~\cite{molinaroSubjectIndependentBiologicalHip2022,arrueLowRankRepresentationHead2020,rappEstimationKinematicsInertial2021,sharifirenaniUseSyntheticIMU2021a}. Japan~\cite{hernandezConvolutionalRecurrentNeural2020,eguchiInsoleBasedEstimationVertical2019}, Korea~\cite{thAccelerationMagnitudeImpact2020,yuDataAugmentationAddress2023}, Italy~\cite{ravizzaMocapDataInertial2020, KarasonGenerativeDataAugHumanBiomechanics}, and the United Kingdom~\cite{bicerGenerativeDeepLearning2022a, liewStrategiesOptimiseMachine2024} each contributed two publications.

\begin{figure}[!h]
\includegraphics[width=10cm]{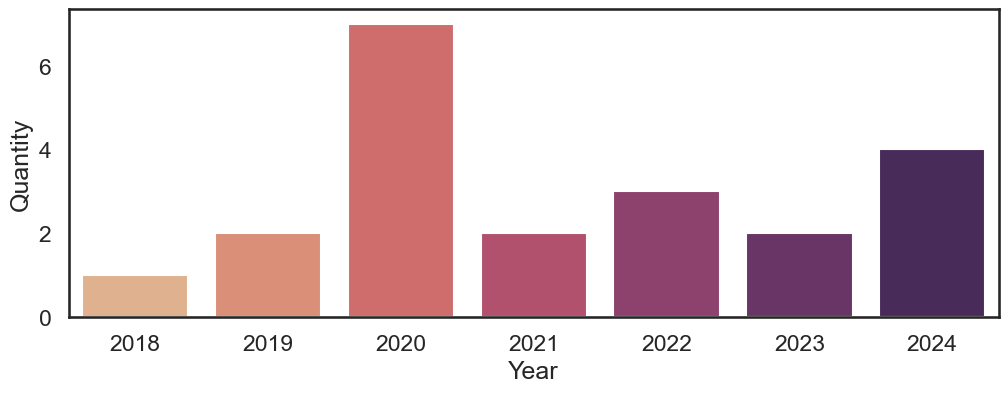}
\caption{{\bf Temporal Trends of Publications}
Number of publications published over the years 2018 to 2024. Notably, the years from 2013 to 2017 are not included as no publications were made that fall within the inclusion criteria.}
\label{fig_temporal}
\end{figure}

An analysis of the temporal distribution of publications reveals interesting trends in research activity over the period considered. As shown in Fig~\ref{fig_temporal}, a notable peak is observed in 2020 with seven publications~\cite{arrueLowRankRepresentationHead2020,mPredictionLowerLimb2020,ravizzaMocapDataInertial2020,mundtEstimationGaitMechanics2020,thAccelerationMagnitudeImpact2020,dorschkyCNNBasedEstimationSagittal2020a,hernandezConvolutionalRecurrentNeural2020}, indicating significant interest in the topic at that time.
This is followed by four publications in 2024~\cite{liangEstimationElectricalMuscle2024,KarasonGenerativeDataAugHumanBiomechanics,tangSyntheticIMUDatasets2024,liewStrategiesOptimiseMachine2024}, reflecting sustained research activity.
In 2022, three publications within this area were published~\cite{bicerGenerativeDeepLearning2022a,molinaroSubjectIndependentBiologicalHip2022,liuSpatiotemporalKinematicCharacteristics2022}. In 2019~\cite{eOptimalControlSimulation2019,eguchiInsoleBasedEstimationVertical2019}, 2021~\cite{sharifirenaniUseSyntheticIMU2021a,rappEstimationKinematicsInertial2021} and 2023~\cite{yuDataAugmentationAddress2023,jProbabilityFusionApproach2023}, two publications each were recorded, indicating consistent research presence during these periods. Conversely, 2018 saw one publication~\cite{zimmermannIMUtosegmentAssignmentOrientation2018}. Notably, no publications were found from 2013 to 2017, suggesting an emergence of interest in the topic in subsequent years.

\subsection*{Reasons for Data Augmentation} \label{Reasons}
The authors of the publications list multiple reasons for using data augmentation for biomechanical time-series data sets: 

\begin{enumerate}
    \item \textbf{Limited Data Set Size}~\cite{mundtEstimationGaitMechanics2020,arrueLowRankRepresentationHead2020,dorschkyCNNBasedEstimationSagittal2020a,eguchiInsoleBasedEstimationVertical2019,rappEstimationKinematicsInertial2021,ravizzaMocapDataInertial2020,bicerGenerativeDeepLearning2022a,sharifirenaniUseSyntheticIMU2021a,liuSpatiotemporalKinematicCharacteristics2022,jProbabilityFusionApproach2023,thAccelerationMagnitudeImpact2020,mPredictionLowerLimb2020,hernandezConvolutionalRecurrentNeural2020, liewStrategiesOptimiseMachine2024}

The main reason for employing data augmentation methods is the limitation in data set size, which impedes the performance of training machine and deep learning models. In this context, the model utilized to solve the actual objective is referred to as a downstream model. Synthetic data is frequently used to improve generalization and prevent overfitting in the downstream model, as reported by several studies~\cite{hernandezConvolutionalRecurrentNeural2020,thAccelerationMagnitudeImpact2020,jProbabilityFusionApproach2023,liuSpatiotemporalKinematicCharacteristics2022,bicerGenerativeDeepLearning2022a,rappEstimationKinematicsInertial2021,eguchiInsoleBasedEstimationVertical2019}.

    \item \textbf{Limited Variation of Movement and Parameters}~\cite{yuDataAugmentationAddress2023,sharifirenaniUseSyntheticIMU2021a,eguchiInsoleBasedEstimationVertical2019}. 

    When data is sparse within a data set and a limited number of participants is used to acquire data, the number of movements and the variety of their execution is limited. Data augmentation can be used to enhance model performance by introducing a broader range of data variations.

    \item \textbf{Data Acquisition Challenges}~\cite{molinaroSubjectIndependentBiologicalHip2022, tangSyntheticIMUDatasets2024}

Acquiring data can be troublesome due to issues such as sensor malfunctions or errors in the data processing pipeline, resulting in data loss. Data augmentation can aid the data collection process and account for missing data. 

    \item \textbf{Sensitivity of IMU Sensors}~\cite{yuDataAugmentationAddress2023, zimmermannIMUtosegmentAssignmentOrientation2018, tangSyntheticIMUDatasets2024}

Inertial measurement unit (IMU) sensors are highly sensitive to the positioning on the body and the rotation they are in. Slight modifications in sensor placement and rotation result in different signals collected from the IMU sensors and thereby hinder a good performance of the model. Data augmentation can help increase the variability in the signals such that the model can adapt to changes in position and rotation. Here,~\cite{zimmermannIMUtosegmentAssignmentOrientation2018} focused on evaluating and explaining the existing gap between synthetic and measured data, related to the influence of soft tissue. 

    \item  \textbf{Financial Constraints}~\cite{eOptimalControlSimulation2019, yuDataAugmentationAddress2023,bicerGenerativeDeepLearning2022a,ravizzaMocapDataInertial2020, liangEstimationElectricalMuscle2024, KarasonGenerativeDataAugHumanBiomechanics, tangSyntheticIMUDatasets2024}

Collecting data and testing new inventions in sports equipment is time and money-consuming. A lot of resources need to be spent on imitatively testing new development of sports equipment, (e.g.\ footwear). Acquiring data from a large number of participants is expensive in terms of money and resources, such as equipment, expert knowledge ,and wages of employees. Data augmentation can help limit these costs and resources due to introducing variability in the data set and thus limit the amount of data acquisition that needs to take place.

    \item  \textbf{Unavailability of Combined Data Set}~\cite{ravizzaMocapDataInertial2020}

There may not be a data set that includes several desired data sources for specific environments. For example, the absence of a Motion Capture (MOCAP) data set in combination with IMU data in microgravity, together with the huge cost of acquiring it, necessitated its synthetic generation.~\cite{ravizzaMocapDataInertial2020}
\end{enumerate}

Hence, the reasons for employing a data augmentation method predominantly either revolve around addressing challenges related to limited data set availability such as restricted movement variations and lower downstream model performance, or focus on the underlying causes for the data limitations, such as expensive data acquisition and sensor issues during data collection. In summary, the reasons listed by the authors demonstrate that data augmentation serves as a versatile and indispensable tool for addressing various challenges in sports product development and related research domains. It enables enhanced model performance, robustness, and generalization capabilities despite data limitations and resource constraints.

\subsection*{Disciplines and Aims}
The publications included explore a diverse range of applications and sports disciplines where data augmentation methods are used for biomechanical time-series data sets. The scoping review reveals a predominant focus on walking tasks, with studies spanning various terrains and surfaces. Additionally, research extends to fall detection, diverse sports disciplines, and specialized areas such as sign language recognition and microgravity training. Each study employs data augmentation techniques to enhance model performance and overcome data limitations, reflecting its crucial role in advancing research across these domains.

\subsubsection*{Walking}
Most publications primarily focused on different types of walking tasks. Even though these publications all focused on increasing the data set size and thereby improving the downstream model's ability to generalize more efficiently, the aims of each publication cover a wide range of use cases:
\begin{itemize}
    \item Creating virtual IMUs from optoelectronic marker trajectories~\cite{mPredictionLowerLimb2020}
    \item Detecting chronic ankle instability~\cite{liuSpatiotemporalKinematicCharacteristics2022}
    \item Analyze errors caused by rotation and misalignment of IMU sensory~\cite{zimmermannIMUtosegmentAssignmentOrientation2018}
    \item Estimation of gait parameters~\cite{rappEstimationKinematicsInertial2021,sharifirenaniUseSyntheticIMU2021a,mundtEstimationGaitMechanics2020}
    \item Improve exoskeleton controls~\cite{molinaroSubjectIndependentBiologicalHip2022,jProbabilityFusionApproach2023}
    \item Reducing resources needed for experiments~\cite{eguchiInsoleBasedEstimationVertical2019}
    \item Estimation of myoelectric activity~\cite{liangEstimationElectricalMuscle2024}
    \item Development of a novel data augmentation technique~\cite{bicerGenerativeDeepLearning2022a,KarasonGenerativeDataAugHumanBiomechanics}
\end{itemize}

\subsubsection*{Diverse Disciplines}
The remaining publications covered a diverse range of sports and disciplines related to human movement biomechanics:

\begin{itemize}
    \item Fall detection applications that aim to predict and categorize falls~\cite{yuDataAugmentationAddress2023,thAccelerationMagnitudeImpact2020, tangSyntheticIMUDatasets2024}
    \item Head impact in contact sports (i.e.\ American football) to predict head injuries~\cite{arrueLowRankRepresentationHead2020}
    \item Running with a focus on the mechanical properties of footwear~\cite{eOptimalControlSimulation2019}
    \item Training in microgravity with the aim of creating a data set for categorizing correctly and incorrectly performed exercises in microgravity~\cite{ravizzaMocapDataInertial2020}
    \item Sign language and the recognition of different signs~\cite{hernandezConvolutionalRecurrentNeural2020}
    \item Running and walking with the aim of estimating sagittal lower body kinetics and kinematics~\cite{dorschkyCNNBasedEstimationSagittal2020a}
    \item Walking and jumping to test strategies for optimizing model performance~\cite{liewStrategiesOptimiseMachine2024}
\end{itemize}

\subsection*{Data Sources and Data Set Details}
Table~\hyperref[S2_Table]{S2} shows the data sources used across all publications within the scope of this review. While ten publications~\cite{molinaroSubjectIndependentBiologicalHip2022,sharifirenaniUseSyntheticIMU2021a,hernandezConvolutionalRecurrentNeural2020,eguchiInsoleBasedEstimationVertical2019,thAccelerationMagnitudeImpact2020,yuDataAugmentationAddress2023,jProbabilityFusionApproach2023, liuSpatiotemporalKinematicCharacteristics2022,bicerGenerativeDeepLearning2022a,liangEstimationElectricalMuscle2024} acquired custom data sets within their publication, seven publications~\cite{eOptimalControlSimulation2019,mPredictionLowerLimb2020,dorschkyCNNBasedEstimationSagittal2020a,arrueLowRankRepresentationHead2020,rappEstimationKinematicsInertial2021, liewStrategiesOptimiseMachine2024,KarasonGenerativeDataAugHumanBiomechanics} used previously collected or public data. The remaining four publications~\cite{zimmermannIMUtosegmentAssignmentOrientation2018,mundtEstimationGaitMechanics2020,ravizzaMocapDataInertial2020,tangSyntheticIMUDatasets2024} used previously collected or public data sets in combination with custom data. 

\subsubsection*{Data Sources}
\begin{figure}[!h]
\includegraphics[width=12cm]{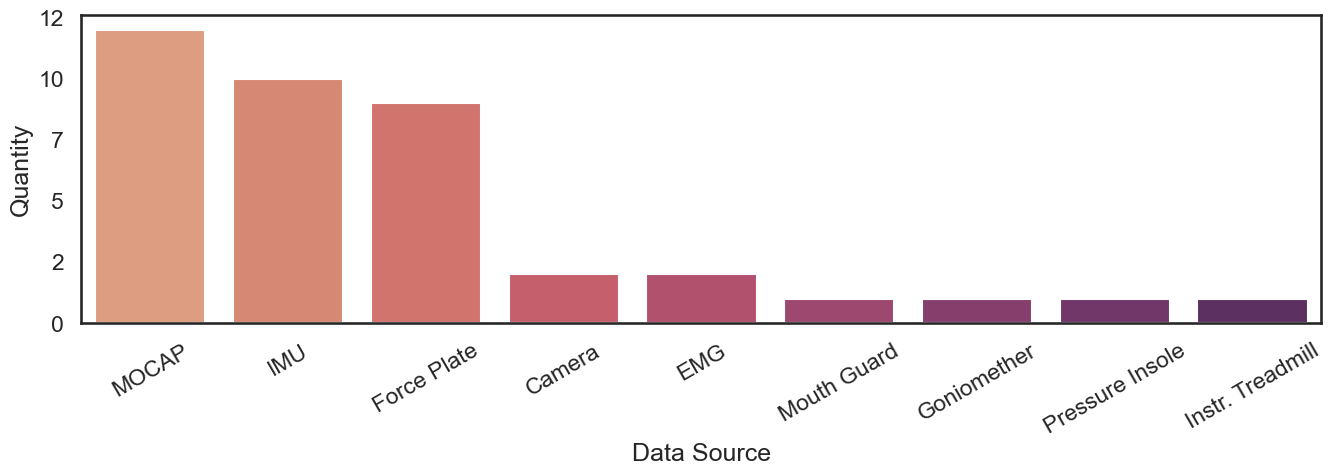}
\caption{{\bf Quantity of Data Sources}
Frequency with which each sensor was used across the publications included in this scoping review. }
\label{fig_sensors}
\end{figure}

Most publications~\cite{thAccelerationMagnitudeImpact2020,liuSpatiotemporalKinematicCharacteristics2022,bicerGenerativeDeepLearning2022a,zimmermannIMUtosegmentAssignmentOrientation2018,arrueLowRankRepresentationHead2020,rappEstimationKinematicsInertial2021,hernandezConvolutionalRecurrentNeural2020,yuDataAugmentationAddress2023,liewStrategiesOptimiseMachine2024} used one type of sensor, mostly MOCAP systems or IMUs. Eight publications~\cite{eOptimalControlSimulation2019,eguchiInsoleBasedEstimationVertical2019,mPredictionLowerLimb2020,mundtEstimationGaitMechanics2020,jProbabilityFusionApproach2023,liangEstimationElectricalMuscle2024,tangSyntheticIMUDatasets2024,ravizzaMocapDataInertial2020} used two different kinds of sensors to create their data set with the most common combination being MOCAP and force plates. \cite{molinaroSubjectIndependentBiologicalHip2022, dorschkyCNNBasedEstimationSagittal2020a,sharifirenaniUseSyntheticIMU2021a} were the only publications that used three different types of sensors - IMUs, MOCAP and Force Plates - to acquire their data set, while~\cite{KarasonGenerativeDataAugHumanBiomechanics} used a data set containing data from five sensors. \cite{arrueLowRankRepresentationHead2020} utilized a data set that was collected using specifically designed mouth guards measuring head impact and~\cite{eguchiInsoleBasedEstimationVertical2019} used a Nintendo Wii Balance Board as a cheaper alternative to force plates. Fig~\ref{fig_sensors} shows the quantity of different data sources used across the publications. 

\subsubsection*{Number of Participants}

The number of participants varied considerably across all publications.
\begin{itemize}
    \item For custom data sets, the number of participants ranged from six to 42 ($\text{mean} = 18.4 \pm 11.82$). 
    \item For publications using previously collected or public data sets, the number of participants ranged from ten to 2295 ($\text{mean} = 564.33 \pm 933.65$), with~\cite{mPredictionLowerLimb2020} and~\cite{arrueLowRankRepresentationHead2020} not reporting the total number of participants.
    \item For publications using combined data sets (both newly acquired and previously collected data), the number of participants in the newly acquired data ranged from six to 32 ($\text{mean} = 17.25 \pm 12.42$). The number of participants from public or previously collected data ranged from 17 to 93 ($\text{mean} = 49.67 \pm 40.05$). Note that~\cite{ravizzaMocapDataInertial2020} did not report the number of participants in the added data set.
\end{itemize}

In summary, the variability in the number of participants across studies underscores the challenges of limited data availability. This presents several issues due to limited movements and limited variations in said movements. The small sample sizes in many studies can lead to limited statistical power, making it difficult to generalize findings to a larger population. Small data sets are more susceptible to random variations and may not capture the full spectrum of potential outcomes. With fewer participants, the diversity of movements captured in the data set is restricted. This limitation can affect the training of machine learning models, which require a wide range of input data to learn effectively. A limited variety of movements means the models may not perform well in real-world scenarios where movements can vary significantly. This includes variations in age, anthropometry, and fitness levels. Consequently, models trained on these data sets may have biased performance and may not apply to all user groups. Therefore, data augmentation emerges as a crucial tool to overcome these challenges as it can help to introduce more variety in the data thereby enabling the development of more effective and generalizable models in the field of biomechanics.

\subsubsection*{Data Set Size}
Only eight publications reported the size of their final data set. However, the size of training instances is hardly comparable and may depend on the data representation.
\begin{itemize}
    \item \cite{dorschkyCNNBasedEstimationSagittal2020a} 595 walking and running cycles
    \item \cite{arrueLowRankRepresentationHead2020} 572 head impact kinematics
    \item \cite{hernandezConvolutionalRecurrentNeural2020} 16890 signs
    \item \cite{thAccelerationMagnitudeImpact2020} reported a total of 1278 samples, however, it was not clear what these samples were. 
    \item \cite{liewStrategiesOptimiseMachine2024} 75732 bilateral trials 
    \item \cite{tangSyntheticIMUDatasets2024} 3313 fall samples
    \item \cite{liangEstimationElectricalMuscle2024} 215 slow, 241 normal, and 258 fast gait cycles
    \item \cite{KarasonGenerativeDataAugHumanBiomechanics} 37687 gait cycles
\end{itemize}

\subsection*{Augmentation Method}
The augmentation methods reported in the publications can be categorized into three main groups: physics-based, classical, and data-driven methods. The following sections will analyze each group regarding their usage and characteristics.

\subsubsection*{Physics-Based Methods}
We categorized ten methods as physical-based methods. These methods either augment data by rotating IMU data to create more IMU signals or use musculoskeletal (MS) models to generate new data samples.  
These techniques mostly ensure the biomechanical validity of the generated samples, but validity may be limited due to simplifications in the formulas used or the realism of the rotations introduced. 
Physics-based methods can be distinguished into the following two categories:

\paragraph{Rotation Methods} 
Rotation methods do not create new synthetic participants but enhance the performance of downstream models by creating variations in sensor signals. In this way, these methods increase the variability in movements and reduce the influence of misaligned sensor data on the performance of downstream models. In four publications, rotation methods are employed to augment the training data and increase the data set size. 

\begin{itemize}
    \item \cite{yuDataAugmentationAddress2023} proposed two rotation-based methods, uniform and normal augmentation, sampling rotation angles from a uniform or truncated normal distribution. 
    \item \cite{rappEstimationKinematicsInertial2021} drew random rotations from a normal distribution around the true orientation of the sensor. Added gravity and random Gaussian noise to account for noise experienced by true sensors.
    \item \cite{mPredictionLowerLimb2020} randomly rotated the relative orientation of virtual sensors to simulate acceleration and gyroscope data. To obtain the angular velocity, they calculated the second derivative of the body segment origin from an MS model.
    \item \cite{zimmermannIMUtosegmentAssignmentOrientation2018} randomly sampled rotation angles and added Gaussian noise to generate synthetic IMU data.
\end{itemize}

\paragraph{Musculoskeletal Models}
The remaining six publications utilized MS models to generate new data, ensuring biomechanical plausibility through anthropometric scaling and fitting of validated body models with appropriate degrees of freedom.

\begin{itemize}
    \item \cite{eOptimalControlSimulation2019} created MS models to solve 1120 optimal control problems using their proposed method.
    \item \cite{dorschkyCNNBasedEstimationSagittal2020a} randomly drew measured joint angles, GRFs, and speeds from training data to generate unique MS models, solving optimal control problems to generate training samples, where simulated data was chosen randomly from 1000 simulations of each subject.
    \item \cite{ravizzaMocapDataInertial2020} built an MS model using OpenSim~\cite{delp2007opensim} and added 3D modeled geometry to simulate microgravity.
    \item \cite{molinaroSubjectIndependentBiologicalHip2022} simulated IMU data using a OpenSim model~\cite{delp2007opensim} due to saturation and dropout of the sensors.
    \item \cite{tangSyntheticIMUDatasets2024} build an openSim model based on markerless MOCAP and augmented data by scaling the simulation model.
    \item \cite{mundtEstimationGaitMechanics2020}
 Set up an anatomical coordinate system based on a biomechanical model, rotated and translated it to match possible sensor positions, and calculated derivatives to obtain acceleration and angular rates.

\end{itemize}

\subsubsection*{Classic Methods}
Six publications used classical methods like adding noise or jittering. While these techniques augment original samples, they do not guarantee biomechanical validity nor add meaningful information of biomechanical patterns, but increase robustness and generalization by adding variability to sensor data.

\begin{itemize}
    \item \cite{eguchiInsoleBasedEstimationVertical2019} proposed probabilistic augmentation, generating an arbitrary number of new steps data based on insole sensor pressure data of one step by drawing from a multivariate normal distribution.
    \item \cite{jProbabilityFusionApproach2023} applied multi-window sampling, scanning input data with a shifting window to provide more data samples.
    \item \cite{hernandezConvolutionalRecurrentNeural2020} applied time and magnitude warping techniques, slicing input into equal lengths and applying distortions and noise.
    \item \cite{thAccelerationMagnitudeImpact2020} compared time warping, jittering, and scaling techniques to change temporal characteristics, scale data magnitude, and add mechanical noise.
    \item \cite{sharifirenaniUseSyntheticIMU2021a} compared various warping techniques.
    \item \cite{liewStrategiesOptimiseMachine2024} applied jittering, magnitude and random guided warping, window slicing and a spawner. 
\end{itemize}

\subsubsection*{Data-Driven Methods}
Five publications used data-driven methods, which are characterized by learning the inherent data distributions and patterns to generate realistic yet synthetic data instances.

\begin{itemize}
\item \cite{liuSpatiotemporalKinematicCharacteristics2022} used a Dual-GAN with gradient penalty to generate spatio-temporal and kinematic data, approximating the real data distribution by a nonlinear dimensionality reduction method (t-SNE algorithm). 
\item \cite{bicerGenerativeDeepLearning2022a} used a GAN with an autoencoder, where the generator (autoencoder) compressed and reconstructed input data, and the discriminator distinguished real from simulated data, creating samples following the training distribution.
\item \cite{arrueLowRankRepresentationHead2020} employed PCA for dimensionality reduction, followed by an emulator to generate new time series, creating a stochastic dimensionality reduction with time-dependent modes.
\item \cite{KarasonGenerativeDataAugHumanBiomechanics} used a Wasserstein GAN with gradient penalty to generate synthetic movement patterns from only anthropometric measures.
\item \cite{liangEstimationElectricalMuscle2024} used a Wasserstein GAN based on the TimeGAN framework to produce synthetic IMU and EMG data.
\end{itemize}

\begin{figure}[!h]
\centering\includegraphics[width=6cm]{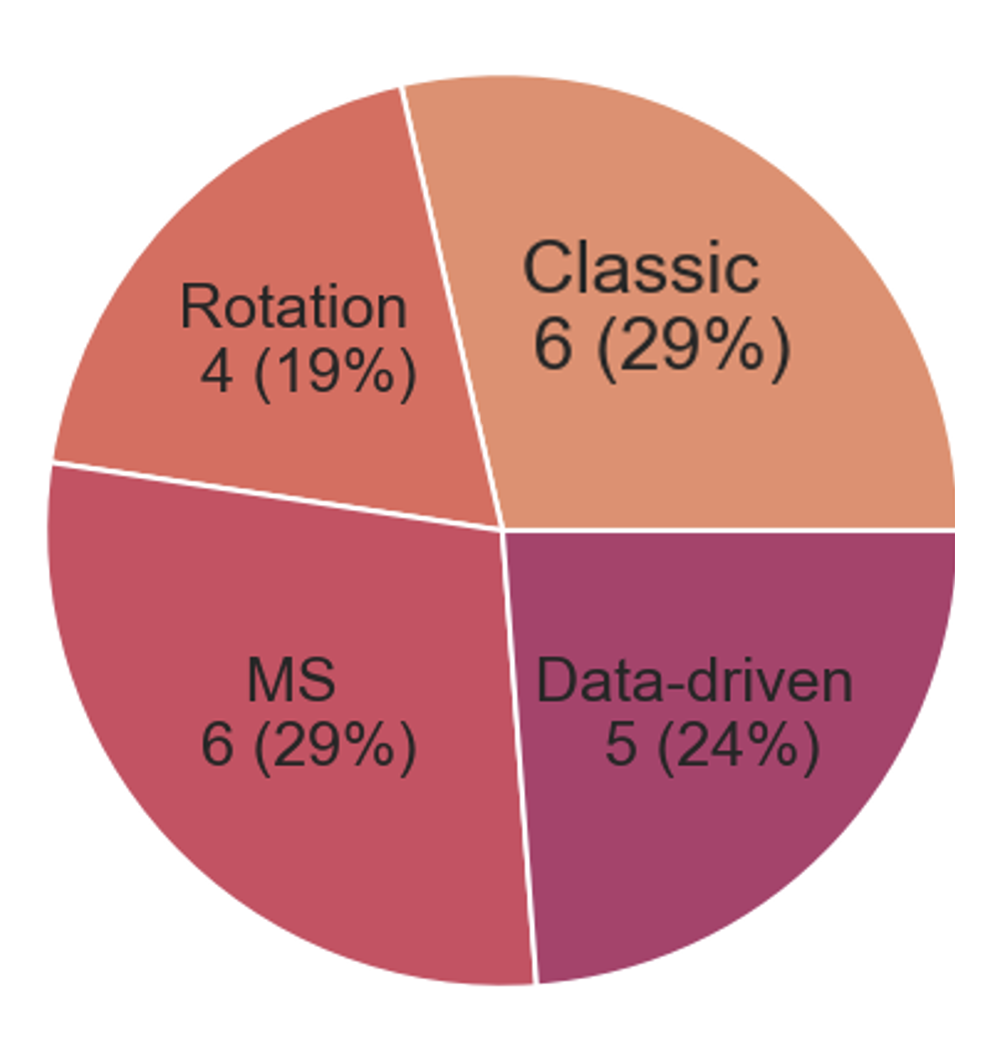}
\caption{{\bf Data Augmentation Methods}
Frequency of each data augmentation category used across the publications included in this scoping review.}
\label{fig_dataAug}
\end{figure}

Fig~\ref{fig_dataAug} illustrates that the various augmentation methods are fairly evenly distributed across the publications. This distribution indicates that there is no single, universally accepted standard method for data augmentation in the field of biomechanical data analysis. Instead, the choice of augmentation method appears to be highly dependent on the specific aims and requirements of each study. Overall, the selection of an augmentation method should consider the specific study goals, the available computational resources, and the desired balance between simplicity and biomechanical fidelity.

\paragraph{Biomechanical Validity} If the biomechanical validity of the generated data is crucial, physics-based methods are superior due to adhering to realistic biomechanical constraints. Additionally, these techniques can address and mitigate the impact of misaligned sensor data, enhancing the performance of the downstream model.
\paragraph{Robustness and Limited Resources} If the aim is to improve the robustness of downstream models without the need for realistic synthetic samples and under limited time resources, classic methods like jittering or warping might be the right choice to increase data variability.
\paragraph{Variability and Biomechanical Patterns} If the aim is to increase data variability and add underlying biomechanical patterns, data-driven methods can be effective, as these methods can learn intricate relationships within the data. While they follow the distribution of the original data set, they can still produce more realistic synthetic samples that reflect the natural variability presented in the data.

\subsubsection*{Noteworthy Methods}
Some publications, although excluded during the full text analysis in the selection process, offer interesting methods that could be beneficial for developing improved data augmentation techniques. Notably, the integration of Reinforcement Learning (RL) in Musculoskeletal (MS) models is a promising approach for enhancing these methods. This integration primarily aims to improve MS models to enhance their biomechanical validity and understanding of complex movements.
For instance,~\cite{DeVree2021654} implemented a Deep Reinforcement Learning (DRL) algorithm using Proximal Policy Optimization (PPO) combined with reward shaping and imitation learning to simulate the walking patterns of both healthy individuals and users of transfemoral prostheses. The results indicated that the prosthesis model required higher muscle forces, demonstrating the added complexity in achieving a natural gait compared to a healthy leg.
Similarly,~\cite{Su2023601} trained an RL agent to simulate human walking by interacting with an MS model. The agent's actions, based on muscle excitations, were optimized through continuous feedback from the environment, leading to improved walking simulations.
Additionally,~\cite{Nowakowski2021671} integrated bioinspired reward reshaping strategies to enhance the simulation and analysis of human locomotion and falls.

These approaches primarily focus on leveraging RL to enhance the biomechanical validity of MS models, allowing for a better understanding of complex human movements. By incorporating biomechanical principles into reward function design, optimizing action and state space management, and ensuring rigorous validation, these models can also be used for data augmentation and may help improve MS models to generate more realistic synthetic data.

\subsection*{Downstream Model} 
In the publications, various downstream models were employed to predict features and evaluate the data augmentation methods. 
A summary of the number of downstream models used per publication and commonly used downstream models is provided below.

\subsubsection*{Overview of Downstream Model Usage}
\begin{itemize}
    \item \textbf{Single Model Usage}: Ten publications~\cite{yuDataAugmentationAddress2023,thAccelerationMagnitudeImpact2020,jProbabilityFusionApproach2023,sharifirenaniUseSyntheticIMU2021a,bicerGenerativeDeepLearning2022a,dorschkyCNNBasedEstimationSagittal2020a,molinaroSubjectIndependentBiologicalHip2022,mundtEstimationGaitMechanics2020, liewStrategiesOptimiseMachine2024, tangSyntheticIMUDatasets2024} used one type of downstream model.
    \item \textbf{Comparison of Two Models:} Five publications~\cite{mPredictionLowerLimb2020,ravizzaMocapDataInertial2020,rappEstimationKinematicsInertial2021,eguchiInsoleBasedEstimationVertical2019,zimmermannIMUtosegmentAssignmentOrientation2018} compared two types of downstream models against each other.
    \item \textbf{Comparison of Multiple Models}: Three publications compared three~\cite{liuSpatiotemporalKinematicCharacteristics2022}, seven~\cite{hernandezConvolutionalRecurrentNeural2020} and eight~\cite{liangEstimationElectricalMuscle2024} types of downstream models, respectively.
    \item\textbf{Publications Without Downstream Models}: Three publications~\cite{eOptimalControlSimulation2019, arrueLowRankRepresentationHead2020, KarasonGenerativeDataAugHumanBiomechanics} did not use any downstream model.
\end{itemize}

\subsubsection*{Common Downstream Models}
\begin{enumerate}
\item{\textbf{Long Short-Term Memory Networks (LSTMs):}}
\begin{itemize}
    \item \textbf{Standard LSTM:}
    \begin{itemize}
        \item Sign language recognition~\cite{hernandezConvolutionalRecurrentNeural2020}
        \item Detecting falls~\cite{tangSyntheticIMUDatasets2024}
        \item Predicting chronic ankle disease~\cite{liuSpatiotemporalKinematicCharacteristics2022}
        \item Prediction of joint kinematics and/or kinetics of various joints~\cite{mPredictionLowerLimb2020, rappEstimationKinematicsInertial2021}
    \end{itemize}
    \item \textbf{ConvLSTM (Combination of LSTM and CNN):}
    \begin{itemize}
        \item Sign language recognition~\cite{hernandezConvolutionalRecurrentNeural2020}
        \item Pre-impact fall detection~\cite{yuDataAugmentationAddress2023}
        \item Predicting chronic ankle disease~\cite{liuSpatiotemporalKinematicCharacteristics2022}
        \item Evaluating joint training~\cite{zimmermannIMUtosegmentAssignmentOrientation2018}
    \end{itemize}
    \item \textbf{Bi-Directional LSTM:}
    \begin{itemize}
        \item Predicting knee joint angles~\cite{sharifirenaniUseSyntheticIMU2021a}
        \item Pre-impact fall detection~\cite{thAccelerationMagnitudeImpact2020}
        \item Estimation of electrical muscle activity~\cite{liangEstimationElectricalMuscle2024}
    \end{itemize}
    \item \textbf{LSTM-Fully Convolutional Network Variant:} Predicting chronic ankle disease~\cite{liuSpatiotemporalKinematicCharacteristics2022}
\end{itemize}

\item{\textbf{Convolutional Neural Networks (CNNs):}}
\begin{itemize}
    \item \textbf{Standard CNN:} 
    \begin{itemize}
        \item Prediction of joint kinematics and/or kinetics of various joints~\cite{dorschkyCNNBasedEstimationSagittal2020a, rappEstimationKinematicsInertial2021, bicerGenerativeDeepLearning2022a}
        \item Evaluating joint training~\cite{zimmermannIMUtosegmentAssignmentOrientation2018}
        \item Predicting foot placement~\cite{jProbabilityFusionApproach2023}
        \item Estimating ground reaction forces~\cite{bicerGenerativeDeepLearning2022a}
        \item Sign language recognition~\cite{hernandezConvolutionalRecurrentNeural2020}
        \item Estimation of electrical muscle activity~\cite{liangEstimationElectricalMuscle2024}
    \end{itemize}
    \item \textbf{Temporal Convolutional Network (TCN):} Predicting sagittal hip moments~\cite{molinaroSubjectIndependentBiologicalHip2022}
     \item \textbf{Attention-Based Multiple-Input Multiple-Output Conv2D Regression Model:} Estimation of electrical muscle activity~\cite{liangEstimationElectricalMuscle2024}
\end{itemize}
\item{\textbf{Other Machine Learning Models:}}
\begin{itemize}
    \item \textbf{K-Nearest Neighbour:}
    \begin{itemize}
        \item Sign language recognition~\cite{hernandezConvolutionalRecurrentNeural2020}
        \item Estimation of electrical muscle activity~\cite{liangEstimationElectricalMuscle2024}
    \end{itemize} 
    \item \textbf{Random Forest:}
    \begin{itemize}
        \item Sign language recognition~\cite{hernandezConvolutionalRecurrentNeural2020} 
        \item Estimation of electrical muscle activity~\cite{liangEstimationElectricalMuscle2024}
    \end{itemize}
    \item \textbf{Decision Tree:} Estimation of electrical muscle activity~\cite{liangEstimationElectricalMuscle2024}
    \item \textbf{Support Vector Machines (SVMs):} 
    \begin{itemize}
        \item Sign language recognition~\cite{hernandezConvolutionalRecurrentNeural2020}
        \item Categorization of correctly and incorrectly performed movements~\cite{ravizzaMocapDataInertial2020}
        \item Estimation of electrical muscle activity~\cite{liangEstimationElectricalMuscle2024}
    \end{itemize}
    \item \textbf{Multi-Layer Perceptrons (MLPs) - Non-CNN:}
    \begin{itemize}
        \item Sign language recognition~\cite{hernandezConvolutionalRecurrentNeural2020}
        \item Prediction of joint angles and moments of lower limbs~\cite{mPredictionLowerLimb2020, mundtEstimationGaitMechanics2020}
        \item Categorization of correctly and incorrectly performed movements~\cite{ravizzaMocapDataInertial2020}
        \item Estimation of electrical muscle activity~\cite{liangEstimationElectricalMuscle2024}
    \end{itemize}
    \item \textbf{Regression Variants:}
    \begin{itemize}
        \item \textbf{Gaussian Processes and Linear Regression:} Estimating vertical GRFs~\cite{eguchiInsoleBasedEstimationVertical2019} 
        \item \textbf{Multinomial Logistic Regression:} Categorize participants in different groups~\cite{liewStrategiesOptimiseMachine2024} 
       
    \end{itemize}

\end{itemize}
\end{enumerate}

The use of LSTMs, particularly their variants such as convLSTMs, was prevalent in many publications, highlighting their popularity in processing sequential biomechanical data. However, the diversity of models used suggests that no single model is universally superior. This variety underscores the importance of selecting appropriate models based on the specific task and data characteristics.

Moreover, the reliance on these models further emphasizes the necessity of data augmentation techniques, especially when dealing with limited data sets. LSTMs and CNNs, in particular, depend on a large amount of training data to avoid overfitting and to enhance generalization. Without sufficient data, these models can struggle to learn robust features and may perform poorly on unseen data~\cite{shortenSurveyImageData2019,iglesias_data_2023}.

\subsection*{Evaluation of Augmentation Methods}
\subsubsection*{Evaluation Methods}
The majority of publications~\cite{dorschkyCNNBasedEstimationSagittal2020a,eguchiInsoleBasedEstimationVertical2019,thAccelerationMagnitudeImpact2020,bicerGenerativeDeepLearning2022a,mPredictionLowerLimb2020,zimmermannIMUtosegmentAssignmentOrientation2018,jProbabilityFusionApproach2023,rappEstimationKinematicsInertial2021,hernandezConvolutionalRecurrentNeural2020,yuDataAugmentationAddress2023,liewStrategiesOptimiseMachine2024} evaluated the quality of data augmentation methods based on the performance of downstream models. Typically, the performance of downstream models trained on non-augmented data was compared to those trained on augmented data. Only a few works~\cite{KarasonGenerativeDataAugHumanBiomechanics,arrueLowRankRepresentationHead2020,mundtEstimationGaitMechanics2020,tangSyntheticIMUDatasets2024} validated synthetic data against the original data set. This was done either visually~\cite{tangSyntheticIMUDatasets2024,arrueLowRankRepresentationHead2020} or statistically~\cite{KarasonGenerativeDataAugHumanBiomechanics,mundtEstimationGaitMechanics2020}, ensuring the augmented data preserved statistical properties. Three publications~\cite{liuSpatiotemporalKinematicCharacteristics2022,liangEstimationElectricalMuscle2024,sharifirenaniUseSyntheticIMU2021a} used both methods to evaluate their augmentation method. Some publications did not evaluate the performance of the data augmentation method~\cite{eOptimalControlSimulation2019,molinaroSubjectIndependentBiologicalHip2022,ravizzaMocapDataInertial2020}.

\paragraph{Limited Comparability} Focusing primarily on downstream model performance provides a direct measure of how effective data augmentation methods are in improving the predictive capabilities of models. While this is crucial, the following limitations should be considered. The lack of comparison of augmentation methods means it is unclear why certain methods work better than others, especially if the details of the data augmentation are not well understood. Understanding the effect of data created using different augmentation techniques is crucial for comprehending the impact of the increased data set on the downstream model and evaluating the quality of certain aspects of the data augmentation technique. Furthermore, comparing synthetic and measured data can help identify gaps and discrepancies. Recognizing where synthetic data falls short can guide the development of more sophisticated augmentation methods that better capture the complexities of real data.

\subsubsection*{Findings}

As shown in Table~\hyperref[S3_Table]{S3}, the consensus was that data augmentation generally enhances the accuracy and generalization of downstream models across individual tasks.

\paragraph{Kinetics vs Kinematics}

Findings on different effects of augmented data on kinematics and kinetics were reported by~\cite{dorschkyCNNBasedEstimationSagittal2020a, mundtEstimationGaitMechanics2020}. Both found that while predictions for kinematics were increased using synthetic samples created by data augmentation methods, there was no improvement in terms of root mean squared error (RMSE) and correlation for kinetics. \cite{mundtEstimationGaitMechanics2020} further reported that adding noise to the training data did increase the performance on kinetics. As 
~\cite{dorschkyCNNBasedEstimationSagittal2020a} used an MS model to generate new synthetic data, the suggestion was made that this discrepancy is 
due to noisy, oscillating joint moments produced by the musculoskeletal model, which only tracked joint angles and GRFs, but not joint moments.
Additionally, the reference joint moments could be too smooth due to filtering applied to marker and force plate data before computing joint moments, leading to inaccuracies, especially for hip joint moments, as they were estimated using inverse dynamics, which accumulated errors.
\cite{mundtEstimationGaitMechanics2020}, which used a rotation-based augmentation technique, argued that predicting kinematics is more challenging than predicting joint moments due to the closer relationship between acceleration and joint moments and the more complex initial value problem of joint angles. Consequently, kinematics predictions benefit from larger data sets, as demonstrated by improved accuracy seen with measured, combined, and simulated data. Conversely, kinetics predictions improve with additional noise in the input data rather than larger data sets, due to the absence of larger soft tissue movements in simulated data. Soft tissue movements affect the calculation of both joint angles and moments, and simulated IMU data lacks these movements, which is a limitation that will be discussed in the next paragraph. \cite{bicerGenerativeDeepLearning2022a} on the other hand reported an improvement in both, kinematic and kinetic predictions, by sampling synthetic data of markers and GRFs simultaneously. This approach demonstrated enhanced test set performances in both kinematic and kinetic predictions, highlighting the potential benefits of more comprehensive data augmentation strategies. 
\paragraph{Soft Tissue Artefacts}

Five publications highlighted issues related to soft tissue artifacts in synthetic data introduced by IMUs attached to various body segments. Synthetic data often lacks typical soft tissue movement artifacts present in measured data, leading to discrepancies known as the “synthetic gap”~\cite{mPredictionLowerLimb2020, dorschkyCNNBasedEstimationSagittal2020a, mundtEstimationGaitMechanics2020,sharifirenaniUseSyntheticIMU2021a,zimmermannIMUtosegmentAssignmentOrientation2018}. This gap significantly influences sensor measurements, especially at higher movement velocities. For instance, determining the hip joint center is challenging due to soft tissue, and rigidly attaching the pelvis IMU is problematic, causing larger artifacts in pelvis measurements~\cite{mPredictionLowerLimb2020, sharifirenaniUseSyntheticIMU2021a}. Some studies suggested that noise from soft tissue movements could improve kinetic predictions, whereas kinematic predictions benefited more from larger data sets~\cite{mundtEstimationGaitMechanics2020}. Additionally, the assumption of rigid sensor attachment to the skeleton when generating synthetic IMU data from MS models does not reflect the loose connections caused by soft tissue in reality, which could be addressed with more realistic models such as wobbling mass models~\cite{dorschkyCNNBasedEstimationSagittal2020a}. Overall, integrating soft tissue artifacts into data augmentation models is crucial for accurately representing measured signals and improving the performance of neural networks in biomechanical analysis~\cite{zimmermannIMUtosegmentAssignmentOrientation2018}.
Most of these publications employed rotation-based augmentation methods, which could be particularly susceptible to soft tissue errors~\cite{mPredictionLowerLimb2020,zimmermannIMUtosegmentAssignmentOrientation2018,mundtEstimationGaitMechanics2020}. The other two publications used an MS model~\cite{dorschkyCNNBasedEstimationSagittal2020a} and different warping techniques~\cite{sharifirenaniUseSyntheticIMU2021a}, respectively. This further emphasizes that soft tissue artifacts might be an issue for several types of data augmentation techniques and therefore need to be considered in the development of more effective methods. On the other hand~\cite{liangEstimationElectricalMuscle2024} offers an approach where the positioning and orientation of IMUs do not need to match exactly, as long as the IMUs are approximately attached to, in their case, thigh, and shank. Even though, they do not address soft tissues within their publication, it would be interesting to know whether they had also issues related to this topic.
In conclusion, synthetic IMU data should be used cautiously when augmenting data sets, as they tend to be too clean and thus not realistic. Additionally, further research is needed to evaluate how soft tissue artifacts can be effectively simulated to ensure a more accurate representation of real-world data.

\paragraph{Discrepancies in Simulated IMU Accuracy and Model Performance}
An interesting discrepancy arises when comparing the simulated IMU accuracy and downstream model performance between two studies,~\cite{sharifirenaniUseSyntheticIMU2021a} and~\cite{mundtEstimationGaitMechanics2020}. In~\cite{sharifirenaniUseSyntheticIMU2021a}, the accuracy of the pelvis IMU was notably poorer than that of other segments, especially for rotational velocities. This study also found that including synthetic IMU data significantly improved kinematic predictions for hip and knee degrees of freedom. However, predictions for internal-external rotation of the hip and knee remained less accurate. Conversely,~\cite{mundtEstimationGaitMechanics2020} reported that the pelvis sensor achieved the highest accuracy among all sensors, with better performance than the leg sensors. Their model results showed that joint moment predictions for the hip were particularly accurate, as were joint angle predictions in the sagittal plane. These differences highlight the variability in simulated IMU data accuracy and suggest that the effectiveness of data augmentation techniques can vary significantly based on sensor locations and types of movements analyzed. The contrasting findings underscore the importance of tailoring data augmentation methods to specific biomechanical contexts and rigorously validating synthetic data against measured data to ensure model reliability and accuracy.

\subsubsection*{Future Research}
Only a few works mentioned future research directions on data augmentation in biomechanical studies. However, they indicate that future research should address several key areas to enhance the effectiveness and realism of synthetic data.

\paragraph{Incorporating Subject-Specific Information}
Conditioning neural networks with inputs that define subjects (e.g.\ anthropometry) and trials (e.g.\ gait cycle duration) during training can help overcome current limitations in estimating absolute differential quantities like velocities and accelerations from synthetic data~\cite{bicerGenerativeDeepLearning2022a}. However, it is worth noting that using only anthropometric data, as in~\cite{KarasonGenerativeDataAugHumanBiomechanics}, is not effective for generating synthetic data samples. It may be beneficial to merge these two approaches, incorporating the anthropometric methodology from~\cite{KarasonGenerativeDataAugHumanBiomechanics} into the framework presented by~\cite{bicerGenerativeDeepLearning2022a}.

\paragraph{Modeling Soft Tissue and Motion Artifacts} Developing more sophisticated soft tissue models and general motion artifact compensation strategies is essential to bridge the reality gap between simulated and measured data. Future work should focus on improving these models to better replicate the dynamics of real-world movements~\cite{zimmermannIMUtosegmentAssignmentOrientation2018, dorschkyCNNBasedEstimationSagittal2020a}.

\paragraph{Weighted Augmentation Strategies} Exploring scientific augmentation strategies that give more weight to challenging motions can improve model robustness and performance. This approach can help in better preparing models to handle difficult scenarios~\cite{yuDataAugmentationAddress2023}.

\paragraph{Domain Adaptation and Iterative Training} Employing domain adaptation techniques, such as GANs, to learn mappings between simulated and measured data can reduce discrepancies. Additionally, iterative data generation during training within a closed loop, such as meta-learning algorithms, can dynamically adjust simulator parameters to produce synthetic data that enhances model accuracy~\cite{dorschkyCNNBasedEstimationSagittal2020a}. 

\bigbreak

In conclusion, advancing data augmentation methods by integrating subject-specific details, improving soft tissue models, employing weighted augmentation strategies, and leveraging domain adaptation and iterative training approaches will accelerate the development of more accurate and realistic biomechanical models. By focusing on these areas, future research can significantly improve the quality and applicability of synthetic biomechanical data, ultimately enhancing the performance of predictive models in this field.

\section*{Conclusions}
This work represents the first scoping review on data augmentation and synthetic data generation for time-series data in human movement biomechanics.
It demonstrates the indispensable role of data augmentation methods in addressing limited data set availability in biomechanical time-series data. 

Key reasons for employing data augmentation include overcoming restricted movement variations, enhancing downstream model performance, and mitigating issues related to expensive data acquisition and sensor inaccuracies. By enabling improved model performance, robustness, and generalization, data augmentation is vital for advancing research in sports product development, health, rehabilitation, and related domains.

This review shows a predominant focus on walking tasks, complemented by other areas of interest, including fall detection, various sports, sign language recognition, and microgravity training. This diversity underscores the importance of data augmentation in improving research across different applications. 

Variability in participant numbers highlights the challenges of limited data, which restricts movement variations and model training effectiveness. Data augmentation not only increases the sample size but also introduces more variety into a data set, thereby, facilitating the development of more effective and generalizable models.

Our findings reveal no single, universally accepted standard for data augmentation in biomechanical data analysis. Instead, methods are chosen based on specific study goals, available resources, and the desired balance between simplicity and biomechanical fidelity. Physics-based methods are recommended for ensuring biomechanical validity, classic methods like jittering or warping are suitable for improving robustness with limited resources, and data-driven methods excel in increasing data variability and reducing biases.

The prevalent use of LSTMs and their variants, such as convLSTMs, as downstream models highlights their popularity in processing sequential biomechanical data. These downstream models are commonly used to evaluate the effectiveness of data augmentation techniques. However, the variety of models used suggests the necessity of selecting appropriate downstream models based on the specific task and data characteristics. This reliance on models emphasizes the need for data augmentation techniques, especially to avoid overfitting and enhance generalization with limited data sets.

The evaluation of data augmentation methods suggests that kinematics predictions benefit from larger data sets, while kinetics predictions improve with additional noise due to the absence of soft tissue movements in simulated data. Synthetic IMU data should be used cautiously as it tends to be too clean and overly smoothed. Thus,  further research is needed to evaluate how soft tissue artifacts can be simulated to ensure a more accurate representation of real-world data.

While focusing on downstream model performance is crucial, it also has limitations. The lack of comparison among augmentation methods means it is unclear why certain techniques work better than others. Understanding the effects of different augmentation techniques is essential for evaluating their impact on downstream models and identifying where synthetic data falls short. This understanding will guide the development of more sophisticated methods that better capture the complexities of real data. 

Hence, we conclude that a systematic comparison of data augmentation techniques for biomechanics, which is currently lacking, is an important task for future research. Such comparisons are vital for advancing the field by providing clearer guidance on the most effective strategies across various applications and data characteristics. Moreover, integrating soft tissue artifacts into data augmentation models, rigorously evaluating synthetic data, and tailoring augmentation methods to specific study needs are essential steps for enhancing the accuracy, robustness, and applicability of predictive models in biomechanical research.

\newpage
\section*{Supporting Information} \label{Supporting Information}

\paragraph*{S1 Table.}
\begin{longtable}{>{\raggedright\arraybackslash}m{8mm}|m{86mm} m{25mm}} \caption*{{\bf Included Publications.} All publications that were included in the final selection of this review. \label{S1_Table}} \\ 
\toprule
\multicolumn{1}{>{\centering\arraybackslash}m{8mm}|}{\textbf{}} 
    & \multicolumn{1}{>{\centering\arraybackslash}m{86mm}}{\textbf{Title}} 
    & \multicolumn{1}{>{\centering\arraybackslash}m{25mm}}{\textbf{Author}}\\ \hline
 \midrule
 \endfirsthead
 \toprule
\multicolumn{1}{>{\centering\arraybackslash}m{8mm}|}{\textbf{}} 
    & \multicolumn{1}{>{\centering\arraybackslash}m{86mm}}{\textbf{Title}} 
    & \multicolumn{1}{>{\centering\arraybackslash}m{25mm}}{\textbf{Author}}\\
 \midrule
 \endhead

 \endfoot

 \bottomrule
 \endlastfoot

\textbf{~\cite{eOptimalControlSimulation2019}}  & Optimal Control Simulation Predicts Effects of Midsole Materials on Energy Cost of Running & Eva Dorschky et al.  \\ \hline

\textbf{~\cite{dorschkyCNNBasedEstimationSagittal2020a}} & CNN-Based Estimation of Sagittal Plane Walking and Running Biomechanics From Measured and Simulated Inertial Sensor Data & Eva Dorschky et al. \\ \hline

\textbf{~\cite{eguchiInsoleBasedEstimationVertical2019}} 
 & Insole-Based Estimation of Vertical Ground Reaction Force Using One-Step Learning With Probabilistic Regression and Data Augmentation & Roy Eguchi et al. \\ \hline

\textbf{~\cite{thAccelerationMagnitudeImpact2020}} & Acceleration Magnitude at Impact Following Loss of Balance Can Be Estimated Using Deep Learning Model & Tae Hyong Kim et al. \\ \hline

\textbf{~\cite{liuSpatiotemporalKinematicCharacteristics2022}} 
  & Spatiotemporal and Kinematic Characteristics Augmentation Using Dual-GAN for Ankle Instability Detection & Xin Liu et al. \\ \hline

\textbf{~\cite{bicerGenerativeDeepLearning2022a}} & Generative Deep Learning Applied to Biomechanics: A New Data Augmentation Technique for Motion Capture Dataets & Metin Bicer et al.  \\ \hline

\textbf{~\cite{mPredictionLowerLimb2020}}
 & Prediction of Lower Limb Joint Angles and Moments during Gait Using Artificial Neural Networks & Marion Mundt et al. \\ \hline
 
\textbf{~\cite{zimmermannIMUtosegmentAssignmentOrientation2018}}
& IMU-to-Segment Assignment and Orientation Alignment for the Lower Body Using Deep Learning & Tobias Zimmermann et al.  \\ \hline

\textbf{~\cite{mundtEstimationGaitMechanics2020}}
& Estimation of Gait Mechanics Based on Simulated and Measured IMU Data Using an Artificial Neural Network & Marion Mundt et al.  \\ \hline

\textbf{~\cite{molinaroSubjectIndependentBiologicalHip2022}}
& Subject-Independent, Biological Hip Moment Estimation During Multimodal Overground Ambulation Using Deep Learning & Dean D. Molinaro et al.  \\ \hline

\textbf{~\cite{arrueLowRankRepresentationHead2020}}
& Low-Rank Representation of Head Impact Kinematics: A Data-Driven Emulator & Patricio Arruè et al. \\ \hline

\textbf{~\cite{rappEstimationKinematicsInertial2021}}
& Estimation of Kinematics from Inertial Measurement Units Using a Combined Deep Learning and Optimization Framework & Eric Rapp et al.  \\ \hline

\textbf{~\cite{sharifirenaniUseSyntheticIMU2021a}} 
& The Use of Synthetic IMU Signals in the Training of Deep Learning Models Significantly Improves the Accuracy of Joint Kinematic Predictions & Mohsen Sharifi Renani et al.  \\ \hline

\textbf{~\cite{hernandezConvolutionalRecurrentNeural2020}} & Convolutional and Recurrent Neural Network for Human Activity Recognition: Application on American Sign Language & Vincent Hernandez et al.  \\ \hline

\textbf{~\cite{yuDataAugmentationAddress2023}} 
& Data Augmentation to Address Various Rotation Errors of Wearable Sensors for Robust Pre-impact Fall Detection & Xiaoqun Yu et al.  \\ \hline

\textbf{~\cite{jProbabilityFusionApproach2023}} 
& A Probability Fusion Approach for Foot Placement Prediction in Complex Terrains & Jingfeng Xiong et al.  \\ \hline

\textbf{~\cite{ravizzaMocapDataInertial2020}}  & From Mocap Data to Inertial Data Through a Biomechanical Model to Classify Countermeasure Exercises Performed on ISS & Martina Ravizza et al.  \\ \hline

\textbf{~\cite{liangEstimationElectricalMuscle2024}}  & Estimation of Electrical Muscle Activity during Gait Using Inertial Measurement Units with Convolution Attention Neural Network and Small-Scale Dataset & Wenqui Liang et al. \\ \hline

\textbf{~\cite{KarasonGenerativeDataAugHumanBiomechanics}}  & Generative Data Augmentation of Human Biomechanics & Halldór Kárason et al.  \\ \hline

\textbf{~\cite{tangSyntheticIMUDatasets2024}}  & Synthetic IMU Datasets and Protocols Can Simplify Fall Detection Experiments and Optimize Sensor Configuration & Jie Tang et al.  \\ \hline

\textbf{~\cite{liewStrategiesOptimiseMachine2024}}  & Strategies to Optimize Machine Learning Classification Performance When Using Biomechanical Features & Bernard Liew et al.  \\ 

\end{longtable}
\newpage
\paragraph*{S2 Table.}
\begin{longtable}{>{\raggedright\arraybackslash}m{10mm}|m{20mm} m{89mm}} 
\caption*{{\bf Data Sources and Data Set Details.} Data sources used in the individual publications, such as Motion Capture (MOCAP), inertial measurement units (IMUs), and electromyography (EMG). Additionally, the details of the data set created using the data sources is listed for each publication. No asterisk (*) denoted publications that used only their custom data set, acquired within the scope of the corresponding publication.  * marks all publications that used previously recorded or public data sets. ** marks publications using a combination of own data and previously recorded or public data. Numbers with exponent $c$ were corrected based on the reference and participant details provided.   \label{S2_Table}} \\ 

 \toprule
\multicolumn{1}{>{\centering\arraybackslash}m{10mm}|}{\textbf{}} 
    & \multicolumn{1}{>{\centering\arraybackslash}m{20mm}}{\textbf{Data Sources}} 
    & \multicolumn{1}{>{\centering\arraybackslash}m{89mm}}{\textbf{Data Set Details}}\\  \hline
 \midrule
 \endfirsthead
 \toprule
\multicolumn{1}{>{\centering\arraybackslash}m{10mm}|}{\textbf{}} 
    & \multicolumn{1}{>{\centering\arraybackslash}m{20mm}}{\textbf{Data Sources}} 
    & \multicolumn{1}{>{\centering\arraybackslash}m{89mm}}{\textbf{Data Set Details}}\\
 \midrule
 \endhead
 \toprule
\multicolumn{1}{>{\centering\arraybackslash}m{10mm}|}{\textbf{}} 
    & \multicolumn{1}{>{\centering\arraybackslash}m{20mm}}{\textbf{Data Sources}} 
    & \multicolumn{1}{>{\centering\arraybackslash}m{89mm}}{\textbf{Data Set Details}}\\
 \midrule
 \endhead
 \endfoot

 \bottomrule
 \endlastfoot
  
\centering\textbf{~\cite{eOptimalControlSimulation2019}}*  &\centering{MOCAP} \\ \centering{Instrumented Treadmill} & Used a public data set from~\cite{FukuchiPublicData2017} containing 28 participants running at different speeds. \\ \hline
\centering\textbf{~\cite{mPredictionLowerLimb2020}}* & \centering{MOCAP} \\ \centering{Force Plate} & Contains data from multiple studies conducted at the German Sport University in Cologne and included trials of participants walking at self-selected speeds. Some of the participants were knee arthroplasty patients. \\ \hline
\centering\textbf{~\cite{zimmermannIMUtosegmentAssignmentOrientation2018}}** & \centering{MOCAP} \\ \centering{IMUs (pelvis, upper and lower leg, foot)} & Used own recordings as well as data from publicly available MOCAP data sets. Data set A: Simulated IMU data from 42 participants performing different walking styles~\cite{CarnegieMellonUniversity}. B and C: Collected real IMU data of four male participants walking back and forth for one minute and 28 participants (13m, 15f) walking six minutes in an eight shape. The data sets B and C are not available anymore using the reference given in their publication.\\ \hline
\centering\textbf{~\cite{dorschkyCNNBasedEstimationSagittal2020a}}* & \centering{MOCAP} \\ \centering{Force Plate} \\ \centering{IMUs (lower back, right thigh, right shank, right foot} & The authors used data previously collected in~\cite{dorschkyEstimationGaitKinematics2019} including 595 walking and running cycles in total, performed by 10 male participants walking and running at six different speeds with 10 trials each. \\ \hline
\centering\textbf{~\cite{mundtEstimationGaitMechanics2020}}** & \centering{MOCAP} \\ \centering{Force Plate} &  The validation set recorded in this publication contained 30 healthy participants (18m, 12f) walking 10 level trials at five different speeds on a 5m walkway. The data of seven participants was excluded due to connection issues and data loss. Additionally, a previously collected data set containing 93 participants (55m, 38f) was used for training~\cite{KomnikData}.\\ \hline
\centering\textbf{~\cite{molinaroSubjectIndependentBiologicalHip2022}} & \centering{MOCAP} \\ \centering{Force Plate} \\ \centering{IMU (trunk, right thigh}  & 16 participants (10m, 6f) completed 10 circles of level ground walking in three different speeds each. Additionally, they completed 10 trials of ramp ascent/descent and stair ascent/descent for different slopes and stair height. IMU data was discarded due to data dropout and was replaced by synthetic data  \\ \hline
\centering\textbf{~\cite{arrueLowRankRepresentationHead2020}}* & \centering{Mouthgard designed in~\cite{Hernandez2015}} & A previous collected data set of 573 head impact kinematic measurement during contact sports was used, containing 6 degrees of freedom kinematics for each impact~\cite{Hernandez2015, LaksariInsights2018}.\\ \hline
\centering\textbf{~\cite{rappEstimationKinematicsInertial2021}}* & \centering{MOCAP} & Participants performed 60-seconds walking and running tasks with self-selected speeds on a treadmill. While some participants were pain-free, others suffered a lower extremity running-related injury. Walking trials from 420 participants (203m, 217f) and running trials from 580 (292m, 288f) were used for the studies. \\ \hline
\centering\textbf{~\cite{sharifirenaniUseSyntheticIMU2021a}} & \centering{MOCAP}\\ \centering{Force Plate} \\ \centering{IMUs (pelvis, left thigh, left shank, left foot)} & $31^c$ participants (12m, 19f), where 13 of them had osteoarthritis, performed 15 trails of 5m walking tasks in three different speeds.\\ \hline
\centering\textbf{~\cite{hernandezConvolutionalRecurrentNeural2020}} & \centering{MOCAP} & 25 male novices in sign language. Each sign (numbers 0 to 10 and 49 words) was taught to them before measurement. In total, 16890 labeled signs were recorded with 60 features representing the kinematics of the right and left sides. \\ \hline
\centering\textbf{~\cite{eguchiInsoleBasedEstimationVertical2019}} & \centering{Pressure Insole} \\ \centering{Force Plate} & Six healthy participants (5m, 1f) walked with a self-selected pace with insole pressure sensors over force plates. They estimated the vGRF based on insole pressure. \\ \hline
\centering\textbf{~\cite{thAccelerationMagnitudeImpact2020}} & \centering{IMU (left side of pelvis)} & 24 participants (14m, 10f) mimic falls occurring among elders in 5 fall directions. In total 1278 falls were recorded. \\\hline
\centering\textbf{~\cite{yuDataAugmentationAddress2023}} & \centering{IMU (lower back)} & The data set used for data augmentation was recorded with 30 participants (15m, 15f) without rotation errors. The validation set was recorded with 12 new participants (6m, 6f) with sensor rotation errors. Each subject performed 21 types of daily activities and 15 falls. All tasks were repeated 5 times, except for static motions that were only repeated once. \\ \hline
\centering\textbf{~\cite{jProbabilityFusionApproach2023}} & \centering{Depth Camera (Waist)} \\ \centering{IMUs (foot, waist)} & 12 participants walked in different settings for 3 min (level ground, ramp ascent/descent) and 10 min (stairs ascent/descent), respectively with three different walking speeds. \\ \hline
\centering\textbf{~\cite{liuSpatiotemporalKinematicCharacteristics2022}} & \centering{MOCAP}  & 30 gait cycles were collected of three male patients diagnosed with CAI walking and 211 normal gait cycles of 10 male control participants with no injuries walking.\\ \hline
\centering\textbf{~\cite{bicerGenerativeDeepLearning2022a}} & \centering{MOCAP}  & Collection of 55 gait cycles across all eight, healthy, participants (4m, 4f) on a horizontal walkway.\\ \hline
\centering\textbf{~\cite{ravizzaMocapDataInertial2020}}** & \centering{MOCAP} \\ \centering{Force Plate}  & Two male athletes and six non astronaut participants (3m, 3f) performed a set of 4 repetitions of squads, wide stance squads and deadlift in correct execution as well as one set of each exercise with wrong execution. Additionally, previously collected experimental data at NASA Johnson Space Center in Houston was used.\\ \hline

\centering\textbf{~\cite{liewStrategiesOptimiseMachine2024}}* & \centering{Force Plate} & In the first data set~\cite{Horsak2020DATA}, 2295 participants (1740 m, 555f) walked at a self-paced speed along 10m across two force plates either barefoot or wearing shoes. A total of 75732 bilateral trials were recorded. The second data set~\cite{Liew2020DATA} contained pressure data of 31 participants (17 healthy and 14 with patellofemoral pain syndrome) performing three trials of CMJ using a self-determined depth. Only GRFs from one side were collected.\\ \hline

\centering\textbf{~\cite{tangSyntheticIMUDatasets2024}}** & \centering{Cameras (right and left front side)} \\ \centering{IMUs (head, chest, waist, front of right thigh and shank, right ankle and right wrist)}  & They collected 16 fall and 16 non-fall actions, each performed three times, of 6 participants. 513 valid samples were collected. Additionally, a public data set~\cite{OEzdDATA} was used containing 20 fall and 16 non-fall actions performed by $14^c$ participants (7m, 7f) repeated between 5 and 6 times. Together 3133 fall samples were provided.  \\ \hline

\centering\textbf{~\cite{liangEstimationElectricalMuscle2024}} & \centering{EMG (biceps femoris, gastrocnemius medialis, gastrocnemius lateralis)} \\ \centering{IMUs (shank, thigh)}  & Seven participants (4m, 3f) walked on a treadmill in a self-selected speed, slow (0.8 times the selected speed) and fast (1.2 times the selected speed). Thereby, data from the sensors was collected for 3 trials for each participant over 40 seconds each. In total, 215 slow, 241 normal, and 258 fast gait cycles were collected. \\ \hline

\centering\textbf{~\cite{KarasonGenerativeDataAugHumanBiomechanics}}* & \centering{EMG} \\ \centering{Force Plate} \\ \centering{IMUs (trunk, thigh, shank, foot)} \\ \centering{Goniomether (Hip, knee, ankle)} \\ \centering{MOCAP} & The authors used a public data set~\cite{camargoDATA} containing walking data from 22 healthy participants (13m, 9f). Hereby, the participants walked in different conditions, including walking on a treadmill, ground level, stairs, ramps at different speeds, stair height, and ramp inclination. A total of 37687 gait cycles is provided. \\

\end{longtable}

\newpage

\paragraph*{S3 Table.}

\begin{longtable}{>{\raggedright\arraybackslash}m{10mm}|m{40mm} m{69mm}} 
\caption*{{\bf Data Augmentation Results.} Reported evaluation methods and results of data augmentation across publications.\label{S3_Table} } \\
 \toprule
\multicolumn{1}{>{\centering\arraybackslash}m{10mm}|}{\textbf{}} 
    & \multicolumn{1}{>{\centering\arraybackslash}m{40mm}}{\textbf{Evaluation}} 
    & \multicolumn{1}{>{\centering\arraybackslash}m{69mm}}{\textbf{Results}}\\ \hline
 \midrule
 \endfirsthead
 \toprule
\multicolumn{1}{>{\centering\arraybackslash}m{10mm}|}{\textbf{}} 
    & \multicolumn{1}{>{\centering\arraybackslash}m{40mm}}{\textbf{Evaluation}} 
    & \multicolumn{1}{>{\centering\arraybackslash}m{69mm}}{\textbf{Results}}\\
 \midrule
 \endhead

 \endfoot

 \bottomrule
 \endlastfoot

\centering\textbf{~\cite{eOptimalControlSimulation2019}} & No evaluation of augmentation method. &\\ \hline
\centering\textbf{~\cite{mPredictionLowerLimb2020}}* & Evaluation of the performance of the downstream model trained on data sets with different numbers of augmented samples & They found that their simulated data corresponds to real data, however, it does take soft tissue into account. Additionally, data augmentation improved the prediction results of the models. However, they did not provide numbers for the improvements achieved by data augmentation. \\ \hline
\centering\textbf{~\cite{zimmermannIMUtosegmentAssignmentOrientation2018}} & Evaluation of synthetic IMU by comparing it to real IMU signals and simulated IMU signals obtained from IMUsim~\cite{young2011imusim}. Comparison of model performance trained on augmented data and non-augmented data based on accuracy. & They found a synthetic gap between real and simulated IMU that could be due to clothing or soft tissue. They reported that recorded data alone is not sufficient, and also training on simulated data alone decreases the performance. However, adding a small amount of recorded IMU data increased the performance up to 92\% accuracy and in summary reported that a combination of simulated and real data yielded promising results. \\ \hline
\centering\textbf{~\cite{dorschkyCNNBasedEstimationSagittal2020a}} & Comparison of model performance trained on augmented data and non-augmented data based on RMSE and Pearson Correlation Coefficient (PCC). & They showed that adding simulated data to the training decreases the RMSE of the joint angles by up to 31\%. PCC improved from 0.96-0.99 to over 0.98 when simulated data is included. However, adding simulated data decreased the performance of hip joint estimates. They also found that simulated data was less noisy than measured data, which can also be related to soft tissue, as mentioned in~\cite{zimmermannIMUtosegmentAssignmentOrientation2018}. It was also reported that overfitting on simulated data could be seen, especially for vertical ground reaction forces.   \\ \hline
\centering\textbf{~\cite{mundtEstimationGaitMechanics2020}} & Comparison between measured and simulated data using the RMSE, correlation coefficient and accuracy. Comparison of model performance trained on augmented data and non-augmented data. &  In their case, the simulated data was a good representation of the measured IMU data, with the pelvis sensor achieving the highest accuracy ($0.95 \pm 0.08$). However, higher gait velocities showed larger derivations between simulated and measured data, which they also attributed to soft tissue movement. Worth mentioning is that they found, that prediction of kinematics increases with more data samples. Introducing additional noise led to better results than increasing the sample size alone.\\ \hline
\centering\textbf{~\cite{molinaroSubjectIndependentBiologicalHip2022}} & No evaluation of augmentation method. &  \\ \hline
\centering\textbf{~\cite{arrueLowRankRepresentationHead2020}} & Visually compared the distribution of the original data set with the distribution of a generated data set containing the same number of samples. &  The synthetic data set had similar distribution as the original data set. Features in the created data set also have no significant differences to the original features, such as peak duration distribution. \\ \hline
\centering\textbf{~\cite{rappEstimationKinematicsInertial2021}} & Comparison of model performance trained on augmented data and non-augmented data based on the RMSE. & They show that models trained on augmented data showed a lower RMSE when tested on augmented data, however, not when tested on non-augmented data. Still, they reported that with a higher number of samples, the RMSE decreases. \\ \hline
\centering\textbf{~\cite{sharifirenaniUseSyntheticIMU2021a}} & Comparison of model performance trained on augmented data and non-augmented data based on the RMSE and correlation coefficient. Comparison of synthetic IMU signals and real IMU signals.&  They report that when trained on simulated and measure data, there is a 54\% reduction in RMSE and a 20\% improvement of the correlation coefficient. The showed an average improvement of RMSE for joint angle predictions of 38\% at the hip and 11\% on the knee when the model was trained only using synthetic data and an improvement of 54\% and 45\% when trained on synthetic and measured data. 
Additionally, they report that the predictions for pelvis signals were worse than predictions from other body locations contradicting~\cite{mundtEstimationGaitMechanics2020}.
\\ \hline
\centering\textbf{~\cite{hernandezConvolutionalRecurrentNeural2020}} & Comparison of model performance trained on augmented data and non-augmented data. & They reported a significant improvement of 3.8\% accuracy when data augmentation was used.   \\ \hline
\centering\textbf{~\cite{eguchiInsoleBasedEstimationVertical2019}} & Comparison of model performance trained on augmented data and non-augmented data based on accuracy, normalized RMSE and normalized error. & Simulating up to three virtual steps for augmenting the data set improved the performance of the models. However, generating more than four steps did not enhance the performance of the model.\\ \hline
\centering\textbf{~\cite{thAccelerationMagnitudeImpact2020}} & Comparison of model performance trained on augmented data and non-augmented data based on the mean absolute percentage error (MAPE). Comparison of different augmentation methods and, additionally, the performance of all augmentation methods applied to one data set. &  The average MAPE values were significantly ($p = 0.01$) decreased by 55.8\% when data augmentation was used. They found no significant difference between the results of the different data augmentation techniques and the results when using all augmentation techniques. \\ \hline
\centering\textbf{~\cite{yuDataAugmentationAddress2023}} & Comparison of model performance trained on augmented data and non-augmented data. & They reported, that the two models trained on augmented data outperformed the model trained on non-augmented data with mean improvements of 6.11\% and 6.5\% when looking at the accuracy on the non-augmented validation set. With increasing improvement when the range of rotation error increased. There was no significant difference between the two different augmentation methods.  \\ \hline
\centering\textbf{~\cite{jProbabilityFusionApproach2023}} & Comparison of model performance trained on augmented data and non-augmented data based on the RMSE and landing feasible area accuracy (LFAA) performance. & Data Augmentation improved the result by 10\% RMSE and 0.74\% LFAA. Stated that data augmentation enhances the prediction accuracy of foot placement. \\ \hline
\centering\textbf{~\cite{liuSpatiotemporalKinematicCharacteristics2022}} & Verification of synthetic IMU based on t-distributed stochastic neighbor embedding. Used different data sets to compare the model performance. Additionally, comparison of model performance trained on augmented data and non-augmented data. & They found that the simulated data followed the distribution of the real data. They reported that the number of training samples is crucial for the models' performance. When adding 100 and 200 synthetic data points, the outcome of the detection model was improved. \\ \hline
\centering\textbf{~\cite{bicerGenerativeDeepLearning2022a}} & Comparison of model performance trained on augmented data and non-augmented data based on accuracy. &  They reported that their data augmentation approach increased the accuracy of estimating joint kinematics by 23\% compared to the model trained on original data only.\\ \hline
\centering\textbf{~\cite{ravizzaMocapDataInertial2020}} & No evaluation of augmentation method. &  \\ \hline

\centering\textbf{~\cite{liewStrategiesOptimiseMachine2024}} & Comparison of model performance trained on augmented data and non-augmented data using multiclass Brier score, area under the curve, balanced accuracy and logarithmic loss  & They reported different performances of the models but did not give a clear evaluation of the augmentation technique. It is reported that for data set A, the best model was trained on using a data set of 8 and 12 times the original size. For data set B, the best model was trained on 2 times the original size when looking at the brier score and 8 and 12 times the original size for the accuracy. \\  \hline

\centering\textbf{~\cite{tangSyntheticIMUDatasets2024}} & The authors illustrated the simulated and real IMU signals and compared the performance when using synthetic data and using experimental data for training. & For non-fall data, the signals of synthetic IMU and measured IMU were similar. For fall, however, the deviation was more significant, which can be due to discrepancy of simulated IMU positions on the model in the simulation environment and the actual IMU placement on subjects. The accuracy improved 4.49\% when trained on simulated instead of experimental data. For the second data set, it is 0.58\% worse than the accuracy achieved when trained on experimental data \\ \hline

\centering\textbf{~\cite{liangEstimationElectricalMuscle2024}} & Comparison of distribution between real and synthetic data using principal component analysis, correlation coefficient, and t-distributed stochastic neighbor embedding. Additionally, comparison of methods trained on data sets containing different amount of synthetic data (1.5 and 2 times training data size). & They found that the distribution of real and synthetic data highly overlap, indicating high similarity. Additionally, the PCC is increased with training size. \\ \hline

\centering\textbf{~\cite{KarasonGenerativeDataAugHumanBiomechanics}} & Compared the synthetic and real ground reaction forces and kinematic coordinate trajectories using statistical parameter mapping (SPM) two-tailed paired t-test ($\alpha = 0.05$). Additionally, t-distributed stochastic neighbor embedding was used to give a qualitative analysis. & It was found that the model is unable to generate synthetic data for unseen subjects. They discard their hypothesis that it is possible to train a generative adversarial network to generate realistic movement by only using anthropometric measures of the subjects, due to the time-series data being subject-dependent. Despite, the SPM had only few non-similarities. 

\end{longtable}

\newpage
\section*{Acknowledgments}
This work was funded by the Austrian Federal Ministry for Climate Action,
Environment, Energy, Mobility, Innovation, and Technology (Project Motion-
Data-Intelligence-Lab; Contract: 2021-0.641.557) and the federal state of
Salzburg within the context of WISS 2025 (Project TexSense; Contract No.
20102-F2101127-FPR).
\nolinenumbers
\newpage
\bibliography{Included}
\end{document}